\documentclass{article}
\usepackage{ijcai22}

\usepackage{times}
\usepackage{soul}
\usepackage{url}
\usepackage[hidelinks]{hyperref}
\usepackage[utf8]{inputenc}
\usepackage[small]{caption}
\usepackage{graphicx}
\usepackage{amsmath}
\usepackage{amsthm}
\usepackage{booktabs}
\usepackage{algorithm}
\usepackage{algorithmic}
\usepackage{multirow}
\usepackage{calc,pifont} 
\usepackage{comment}
\usepackage{amssymb,mathrsfs}
\usepackage{bbding}
\usepackage{arydshln}
\title{Survey on Graph Neural Network Acceleration: An Algorithmic Perspective}

\author{
Xin Liu$^{1,2}$ \and
Mingyu Yan$^1$\footnote{Corresponding author} \and
Lei Deng$^3$ \and
Guoqi Li$^{2,4}$ \and
Xiaochun Ye$^{1,2}$ \and \\
Dongrui Fan$^{1,2}$ \and %\textit{, Senior Member,~IEEE}
Shirui Pan$^{5}$ \And %\textit{, Member,~IEEE}
Yuan Xie$^{6}$\\ %\textit{, Fellow,~IEEE}
\affiliations
$^1$ SKLCA, Institute of Computing Technology, Chinese Academy of Sciences, China\\
$^2$ University of Chinese Academy of Sciences, China\\
$^3$ Tsinghua University, China\\
$^4$ Institute of Automation, Chinese Academy of Sciences, China\\
$^5$ Monash University, Australia\\
$^6$ University of California, Santa Barbara, America\\
\emails
\{liuxin19g, yanmingyu, yexiaochun, fandr\}@ict.ac.cn,
leideng@mail.tsinghua.edu.cn,
guoqi.li@ia.ac.cn,
shirui.pan@monash.edu,
yuanxie@ucsb.edu
}

\begin{document}

\maketitle

\begin{abstract}
    Graph neural networks (GNNs) have been a hot spot of recent research and are widely utilized in diverse applications. However, with the use of huger data and deeper models, an urgent demand is unsurprisingly made to accelerate GNNs for more efficient execution. In this paper, we provide a comprehensive survey on acceleration methods for GNNs from an algorithmic perspective. We first present a new taxonomy to classify existing acceleration methods into five categories. Based on the classification, we systematically discuss these methods and highlight their correlations. Next, we provide comparisons from aspects of the efficiency and characteristics of these methods. Finally, we suggest some promising prospects for future research.
\end{abstract}

\section{Introduction}

Graph Neural Networks (GNNs) \cite{scarselli2008graph} are deep learning based models that apply neural networks to graph learning and representation. They are technically skillful \cite{kipf2016semi,velickovic2018graph} and theoretically supported \cite{pope2019explainability,ying2019gnnexplainer}, holding state-of-the-art performance on diverse graph-related tasks \cite{hamilton2017inductive,xu2018how}. 
Owing to the significant success of GNNs in various applications, recent years have witnessed increasing research interests in GNNs, hastening the emergence of reviews that focus on different research areas. A few reviews \cite{wu2020comprehensive,zhang2020deep,battaglia2018relational} pay close attention to GNN models and generic applications, while others \cite{ijcai-WangHW0SOC0Y21,ijcai-ZhangW021} place emphasis on specific usages of GNNs. Moreover, hardware-related architectures \cite{abadal2021computing,han2021survey} and software-related algorithms \cite{lamb2020graph} of GNNs are also emphatically surveyed by researchers. Thereby, the above reviews further promote the widespread use of GNNs. However, as GNNs are widely used in emerging scenarios, it is discovered that GNNs are plagued by some obstacles that lead to slow execution.
Next, we will discuss obstacles that restrict the efficiency of GNNs execution and present our motivation.

\textbf{Motivation: why GNNs need acceleration?} 

\textit{First}, \textbf{explosive increase of graph data} poses a great challenge to GNN training on large-scale datasets. Previously, many graph-based tasks were often conducted on toy datasets that are relatively small compared to graphs in realistic applications, which is harmful to model scalability and practical usages. Currently, large-scale graph datasets are thereby proposed in literature \cite{OGB} for advanced research, and at the same time, making GNNs execution (i.e., training and inference) a time-consuming process.
\textit{Second}, under the condition that the over-smoothing issue has been skillfully avoided \cite{dropedge1}, using \textbf{deeper and more complicated structures} is a promising way to acquire a GNN model with good ability of expression \cite{ClusterGCN,dropedge1}, which, on the other hand, will increase the time cost of training a well-expressive model.
\textit{Third}, special devices, such as edge devices, generally have \textbf{strict time restrictions on GNN training and inference}, especially in a time-sensitive task. Due to the limited computing and storage resources, the training and inference time on such devices can easily become intolerable. Therefore, it is still an urgent need to accelerate GNNs in both training and inference.

However, no literature has systematically investigated acceleration methods for GNNs at the algorithm level. Practically, algorithm level optimizations not only promote the model accuracy but also accelerate the model learning \cite{chen2018fastgcn,bojchevski2020scaling}. We argue that algorithmic acceleration methods for GNNs will greatly benefit processes of training and inference, and at the same time, the overall performance of GNN frameworks \cite{PyG,DGL}, since a well-designed framework equipped with an optimized algorithm can empirically gain a two-fold promotion \cite{GP_Pagraph}. In consequence, despite some insightful reviews on graph-related frameworks and hardware accelerators, a thorough review on algorithmic acceleration methods for GNNs is highly expected, which is exactly the goal and the focus of this work.

In this paper, we provide a comprehensive survey on algorithmic acceleration methods for GNNs, in which graph-level and model-level optimizations are emphatically focused. To summarize, we highlight our contributions as follows:

1) \textbf{New Taxonomy:} we classify existing methods into five categories via a double-level taxonomy that jointly considers optimized factors and core mechanisms (see Section \ref{sec:2}).

2) \textbf{Comprehensive Review:} we provide a comprehensive survey on existing methods and introduce these methods by categories. And we emphatically focus on common grounds and unique points among these methods (see Section \ref{sec:3}).

3) \textbf{Thorough Comparison:} we summarize the performance of training time of typical acceleration methods and further give a thorough comparison from an overall perspective, in which correlations among these methods are particularly highlighted (see Section \ref{sec:4}).

4) \textbf{Future Prospects:} based on the overall comparison, we discuss some potential prospects of GNNs acceleration for reference (see Section \ref{sec:5}).

%################ Section 2 ################

\section{Preliminary and Taxonomy} \label{sec:2}
In this section, we first introduce the background of GNNs and the conventional execution including processes of training and inference. Then, we propose a taxonomy of acceleration methods for GNNs. Notations and the corresponding descriptions used in the rest of the paper are given in Table \ref{tab:Notation}. 

\tabcolsep 3.pt
\begin{table}[t] %\vspace{-2.2mm}
\centering
\begin{tabular}{cc}
\bottomrule
\textbf{Notations} & \textbf{Descriptions} \\
\bottomrule
\textbf{H} & Hidden feature matrix of graph\\
\textbf{h} & Hidden feature of a node\\
\textbf{X} & Feature matrix of graph\\
\textbf{A} & Original adjacency matrix\\
\textbf{A}$_{sp}$, $\widetilde{\textbf{A}}$ & Sparsified and normalized adjacency matrix\\
\textbf{D}, $\widetilde{\textbf{D}}$ & Degree matrix of \textbf{A} and $\widetilde{\textbf{A}}$\\
\textbf{S} & Normalized adjacency matrix with self-loops\\
\textbf{W} & Weight matrix of graph\\
$\sigma$ & Nonlinear activation function\\
\textit{V}, \textit{E} & Node and edge sets of a graph\\
\textit{N(v)}, \textit{SN(v)} & Original and sampled sets of \textit{v}'s neighbors\\
\bottomrule
\end{tabular}
\vspace{-1.5mm}
\caption{Notations and corresponding descriptions used in the paper.} 
\label{tab:Notation}
\end{table} %\vspace{-4.mm}

\subsection{Background of GNNs and Model Execution}

To efficiently capture hidden patterns in graphs, GNNs provide an inspired idea of combining the design of modern neural networks and graph learning \cite{wu2020comprehensive}. With the background of deep learning, many variants of GNN are proposed by adding particular mechanisms to the original GNN model, e.g., Graph Convolutional Networks (GCNs) \cite{kipf2016semi}, Graph Attention Networks (GATs) \cite{velickovic2018graph}, and Graph Isomorphism Networks (GINs) \cite{xu2018how}. As with most artificial neural networks (ANNs), the execution of GNNs contains processes of training and inference. Herein, we take the GCN model as an exemplar for formulated introduction, owing to its powerful ability and the widespread usage of handling graph-related tasks. Generally, given \textbf{A} and \textbf{X} as input, the forward propagation in the \textit{l}-th layer in GCN training can be formulated as: 
\begin{equation} \label{Eq1} 
    \textbf{H}^{l} = \sigma \left( \widetilde{\textbf{D}}^{-1/2} \widetilde{\textbf{A}} \widetilde{\textbf{D}}^{-1/2} \textbf{H}^{l-1} \textbf{W}^{l-1} \right).
\end{equation}
The backpropagation is performed by updating \textbf{W} via the computed gradient. Distinctly, Equation \ref{Eq1} implies the forward propagation to be an iterative computing process in a layered manner, which takes non-trivial cost in terms of time and storage. As for inference, the trained model is utilized to infer and acquire hidden representation for downstream tasks. %Since the inference process is more commonly to be deployed on resource-limited devices, accelerating model inference is also a critical and urgent need for various applications. 
In consequence, for training, it is time-consuming to obtain the well-trained \textbf{W}; for inference, attempts of deploying the inference to resource-limited devices make an urgent demand for inference efficiency.

\begin{figure}[t] %\vspace{-2mm}
\centering
\includegraphics[width=0.99\columnwidth]{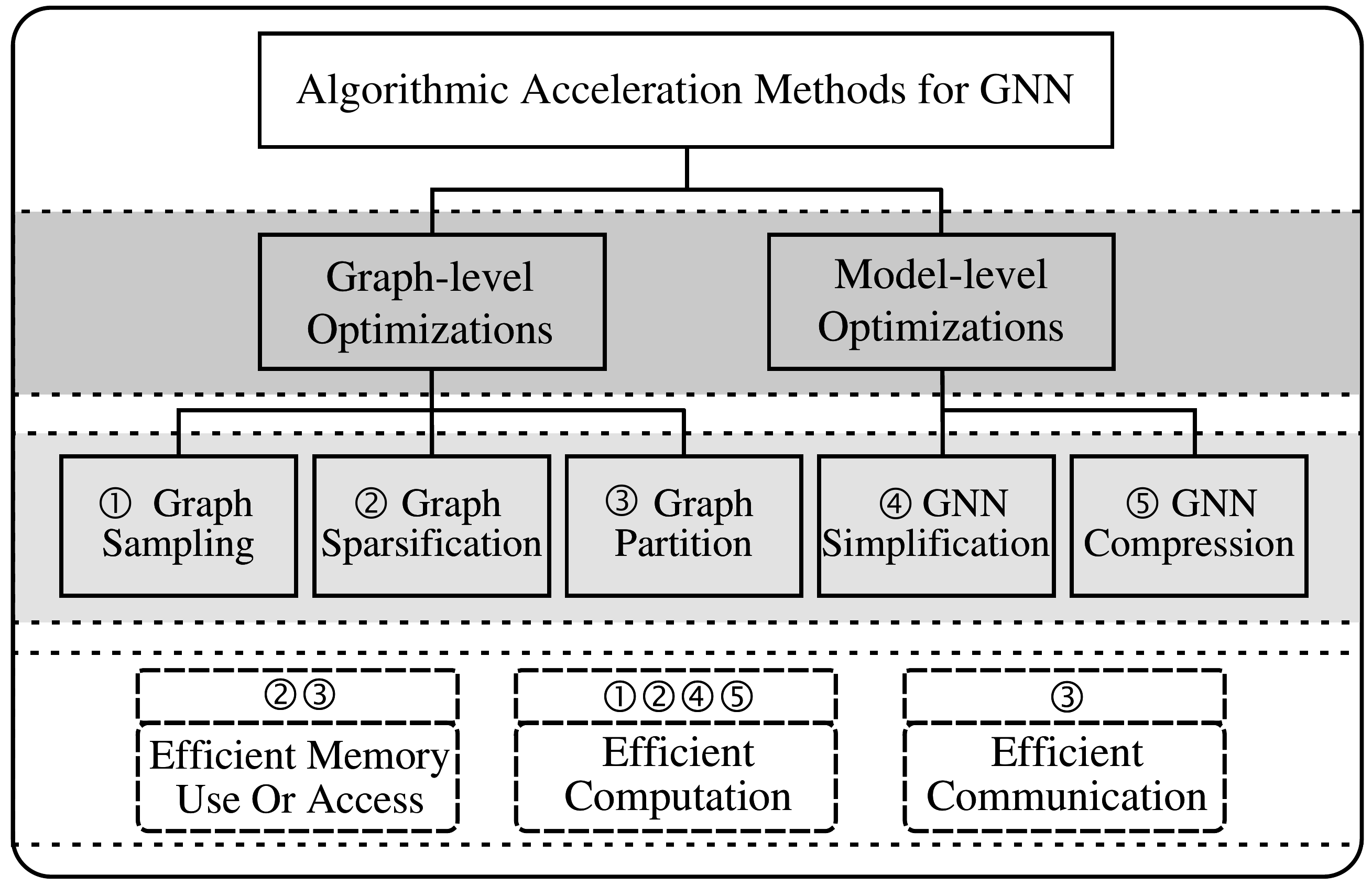}
\vspace{-1.5mm}
\caption{Taxonomy of acceleration methods for GNNs. These methods are classified into five categories by a double-level decision, i.e., the optimized factor in GNN execution (1st level) and the core mechanisms of these methods (2nd level). Moreover, five categories of methods are highlighted by their objectives. For instance, methods of ``Graph Sparsification'' and ``Graph Partition'' can speed up GNN execution by adopting efficient memory access or usage.}
\label{fig:taxonomy}
\end{figure} %\vspace{-5mm}

\subsection{Taxonomy of Acceleration Methods}
A double-level taxonomy of existing algorithmic acceleration methods for GNNs is illustrated in Figure \ref{fig:taxonomy}. First, methods are classified into two major categories according to the optimized factor. ``Graph-level'' denotes that the optimization is conducted on graphs used for training and inference by modifying topology or density of graphs. ``Model-level'' denotes that the optimization is made on GNNs' model containing modifications to the model structure or weight. Further, we divide these methods into five categories based on their mechanisms, i.e., graph sampling, graph sparsification, graph partition, GNN simplification, and GNN compression. As an exemplar, ``Graph Sampling'' denotes that graph sampling methods are utilized to accelerate the training convergence of GNNs. At the final level, we label these methods by their optimization objectives, e.g., graph sampling gains acceleration by reducing the computation cost. Detailed discussions are given in Section \ref{sec:3}.
%e.g., graph sampling algorithms gain acceleration by reducing the computation cost. Detailed discussions are given in section \ref{sec:3}.

%################ Section 3 ################

\section{Acceleration Methods for GNNs} \label{sec:3}
In this section, we discuss five types of algorithmic acceleration methods that respectively focus on graph-level and model-level optimizations. For each category of methods, we first introduce the fundamental and the way for acceleration from an overall aspect. Then, we exemplify typical work belonging to these categories and highlight their correlations.

\subsection{Graph-level Optimizations}

%%%%%%%%%%%%%%%%%%%%%%%%%%%3.1.1%%%%%%%%%%%%%%%%%%%%%%%%%%%
\begin{figure*}[!ht]
\centering
\includegraphics[width=0.72\linewidth]{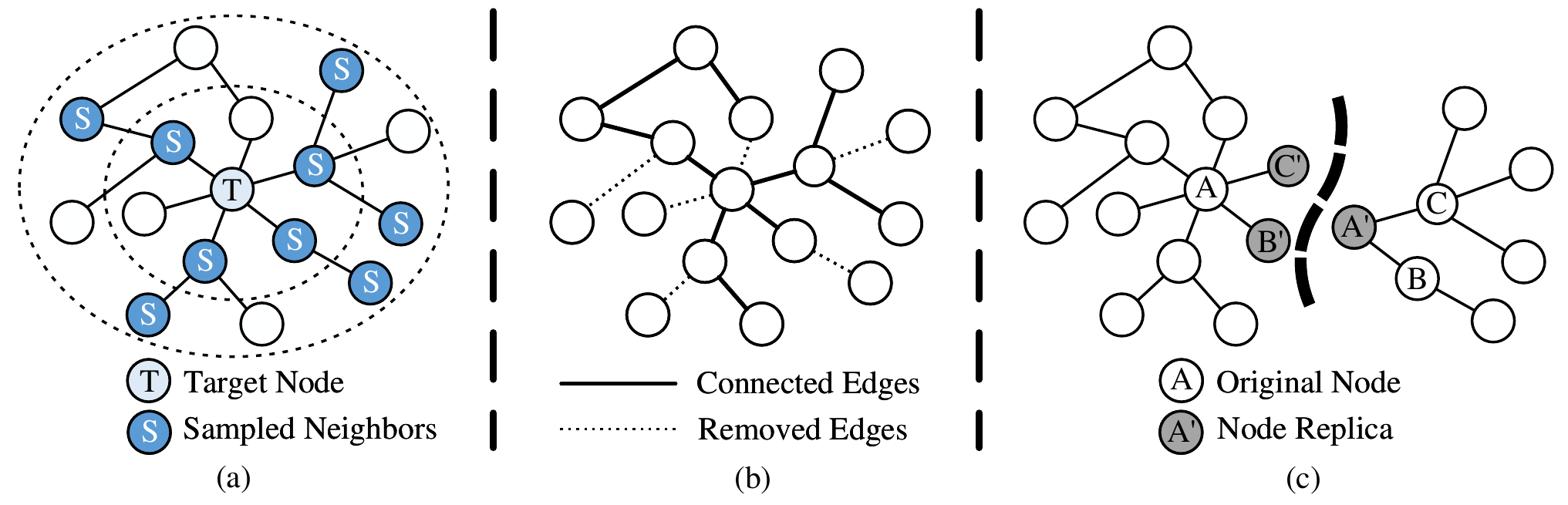}
\vspace{-1.5mm}
\caption{Illustrations of graph-level improvements: (a) a graph sampling method that samples 2-hop neighbors; (b) a graph sparsification method that removes useless edges; (c) a graph partition method that divides a graph into two subgraphs with node replicas preserved.}
\label{fig:graph-level}
\end{figure*} 

\subsubsection{\ding{172} Graph Sampling}

Conventional training of GNNs, especially GCNs, is executed in a full-batch manner, restricting model update to once per epoch and thus slows down the training convergence. As illustrated in Figure \ref{fig:graph-level}(a), graph sampling methods generally select partial nodes in one node's neighborhood or one layer in a graph to acquire subgraphs for subsequent training, which makes \textbf{efficient computation} for model training. We formulate the sampling-based training process of GNNs using GraphSAGE \cite{hamilton2017inductive} as an exemplar:
\begin{equation} \label{eq:sample} 
    SN\left( v \right) = \texttt{Sampling}^{(l)} \left( N\left( v \right) \right)
\end{equation}  \vspace{-4mm}
\begin{equation} \label{eq:aggregate}
    \textbf{h}^{(l)}_{N(v)} = \texttt{Aggregate}^{(l)} ( \lbrace \textbf{h}_{u}^{(l-1)}: u\in SN(v) \rbrace )
\end{equation}  \vspace{-2mm}
\begin{equation} \label{eq:update}
    \textbf{h}^{(l)}_{v} = \texttt{Update}^{(l)} ( \textbf{h}^{(l-1)}_{v} , \textbf{h}^{(l)}_{N(v)} )
\end{equation}
Functions \texttt{Aggregate} and \texttt{Update} denote processes for aggregating hidden features from neighbors and updating features, respectively.
As we can observe, the sampling method restricts size of $N(v)$ to $SN(v)$ and decreases the computation cost in subsequent step, instead of generating \textbf{h} conventionally with a whole graph utilized. Moreover, sampling methods utilize a mini-batch strategy in training and update the model once per batch, which accelerates the model convergence and promotes the training efficiency. 

Please note that, sampling methods for GNN training are generally diverse and are different in design purposes, e.g., for accelerating model training or reducing memory usage. 
%The interested reader will find a more detailed review focused on sampling-based GNNs training in the literature \cite{liu2021sampling}. 
Herein, we pay close attention to methods that benefit the training speed and convergence. 
%Next up, these methods are discussed in order according to a categorization proposed by the literature \cite{liu2021sampling}.
Next up, by referring to the practice of previous literature \cite{liu2021sampling}, we orderly discuss these methods as follows.

\noindent $\bullet$  \textbf{\textit{Node-wise sampling methods:}} node-wise sampling is a fundamental sampling method that focuses on each node and its neighbors in a training graph. GraphSAGE \cite{hamilton2017inductive} is an inductive learning framework in which a sampling method and a mini-batch strategy are first proposed to benefit the training. In the sampling process, GraphSAGE randomly selects 2-hop neighbors for each node in a batched manner. The aggregation process is performed based on the sampled result. Taking inspiration from GraphSAGE, VR-GCN \cite{chen2018stochastic} restricts the number of neighbors per node to an arbitrarily small size for alleviating the exponential growth of the receptive field.

\noindent $\bullet$  \textbf{\textit{Layer-wise sampling methods:}} layer-wise sampling can be regarded as an algorithmic improvement of node-wise sampling. Since node-wise sampling suffers from exponential expansion of multi-hop neighbors, layer-wise sampling alleviates the heavy overhead by sampling a fixed number of nodes in each layer based on pre-computed probability. 
Typically, FastGCN \cite{chen2018fastgcn} independently samples a certain number of nodes per layer and reconstructs connections (of nodes) between two successive layers according to \textbf{A} of the training graph. 
Based on a hierarchical model as well, the sampling process of AS-GCN \cite{huang2018adaptive} is layer-dependent and probability-based. It samples nodes according to the parent nodes sampled in the upper layer.

\noindent $\bullet$  \textbf{\textit{Subgraph-based sampling methods:}} subgraph-based sampling generates subgraphs for training in a two-step manner: sampling nodes and constructing connections (edges). The graph samplers are varied compared to the first two sampling methods. For instance, GraphSAINT \cite{graphsaint-iclr20} utilizes three samplers, i.e., node sampler, edge sampler, and random walk sampler, to sample nodes (or edges) and construct a subgraph in each batch. Moreover, subgraph sampling can be parallelized at the processing unit level with a training scheduler aided \cite{zeng2019accurate}, making the training efficient and easily scalable. Such parallelization takes good advantage of the property that subgraphs can be sampled independently.

\noindent $\bullet$  \textbf{\textit{Heterogeneous sampling methods:}} heterogeneous sampling is designed to accelerate the training and handle the graph heterogeneity, i.e., imbalanced number and type of neighbors in a node's neighborhood. HetGNN \cite{zhang2019heterogeneous} resolves the heterogeneity issue by traversing and collecting neighbors in a random walk manner. The collected neighbors are grouped by type and are further sampled based on the visit frequency. To achieve balanced neighboring distribution, HGSampling is proposed in HGT \cite{hu2020heterogeneous} to sample different types of nodes orderly. HGSampling is probability-based and ensures a balanced sampling result among diverse neighbors.

\vspace{-1mm}

%%%%%%%%%%%%%%%%%%%%%%%%%%%3.1.2%%%%%%%%%%%%%%%%%%%%%%%%%%%
\subsubsection{\ding{173} Graph Sparsification}
Graph sparsification is a classic technique for speeding up many fully dynamic graph algorithms \cite{eppstein1997sparsification}. As illustrated in Figure \ref{fig:graph-level}(b), graph sparsification methods typically remove task-irrelevant edges in a graph by designing a specific optimization goal. Recent graph sparsification methods propose to sparsify input graphs before they are fed into a GNN model, which makes \textbf{efficient computation and memory access} for model training. To be generic, we formulate the sparsification method as follows: %\vspace{-2mm}
\begin{equation} \label{eq:sparse}
    \resizebox{0.91\linewidth}{!}{$
    \textbf{A}_{sp} = \texttt{Sparse}(\textbf{A}) = \left\{
    \begin{aligned}
    \texttt{Sp.Algo.}(\textbf{A})&, \quad \textit{Heuristic} \\
    \texttt{Sparsifier}(\textbf{A})&, \quad \textit{Learnable}
    \end{aligned}
    \right.
    $}
\end{equation} %\vspace{-3.9mm}

\begin{equation}
    \{\textbf{A}_{sp} \rightarrow GNN \} \rightarrow \textit{Training \& Inference} 
\end{equation} 
In Equation \ref{eq:sparse}, the \texttt{Sparse} function can be implemented via two schemes: 1) designing a heuristic sparsification algorithm; 2) building a learnable module (e.g., sparsifier) for sparsification. After processing, the graph ($\textbf{A}_{sp}$) is highly sparse in general, where task-irrelevant edges are removed to reduce subsequent computation and redundant memory access cost in GNN. Moreover, graph sparsification methods can be skillfully leveraged to reduce the communication latency during training, thus accelerating GNN training in the hardware \cite{dropedge3}. Next, we discuss these methods according to their schemes (i.e., heuristic or learnable).

\noindent $\bullet$  \textbf{\textit{Heuristic sparsification:}} DropEdge \cite{dropedge1} proposes to randomly remove edges in each training epoch, which aims to resolve the over-smoothing issue in deep GNN training. Instead of designing a global optimization goal, random edge dropping is heuristic and ensures both fast execution of algorithm and randomness of graph. FastGAT \cite{sparse1} presents a resistance-based spectral graph sparsification solution to remove useless edges, reducing the number of attention coefficients in a GAT model. Thereby, both training and inference are well accelerated.

\noindent $\bullet$  \textbf{\textit{Learnable module for sparsification:}} NeuralSparse \cite{sparse5} and SGCN \cite{sparse3} cast graph sparsification as an optimization problem. NeuralSparse utilizes a deep neural network (DNN) to learn a sparsification strategy based on the feedback of downstream tasks in training. It casts graph sparsification as an approximative objective of generating sparse \textit{k}-neighbor subgraphs. SGCN formulates graph sparsification as an optimization problem and resolves it via an alternating direction method of multipliers (ADMM) approach \cite{ADMM}. %achieving faster convergence compared to DropEdge.
Additionally, GAUG \cite{dropedge2} proposes two variants: GAUG-M utilizes an edge predictor to acquire probabilities of edges in a graph, and modifies input graphs based on the predicted probabilities; GAUG-O integrates the edge predictor and GNN model to jointly promote edge prediction and model accuracy.

Other literature aims to accelerate GNN inference, such as UGS \cite{sparse2} and AdaptiveGCN \cite{sparse4}. UGS utilizes lottery ticket hypothesis to sparsify input graphs and a GNN model iteratively. AdaptiveGCN builds an edge predictor module to remove task-irrelevant edges for inference acceleration on both CPU and GPU platforms.

%%%%%%%%%%%%%%%%%%%%%%%%%%%3.1.3%%%%%%%%%%%%%%%%%%%%%%%%%%%
\subsubsection{\ding{174} Graph Partition}

Graph partition has been an NP-hard problem with the goal of reducing the original graph to smaller subgraphs that are constructed by mutually exclusive node groups \cite{GP_Survey}. As graph data goes enormous, studies have previously deployed GNN execution on diverse platforms, such as a distributed system with multiple GPUs equipped, which places high demands for data communication in systems. 
Thereby, to accelerate GNN execution on such systems, graph partition methods are introduced to reduce the communication cost and maintain the load balance (\textbf{efficient communication and memory usage}). Figure \ref{fig:graph-level}(c) illustrates a simple paradigm of graph partition using a vertex-cut strategy. We generally formulate the process of graph partition as follows: %\vspace{-1.9mm}
\begin{equation}
    \textit{V}_{1} \cup \cdots \cup \textit{V}_{k} = \texttt{Divide}(\textit{V}), \{ \textit{V}_{i} \cap \textit{V}_{j} = \emptyset \quad \forall i \neq j \}
\end{equation} \vspace{-2mm}
\begin{equation}
    \{\textit{Block}_{i}| i \ge 1 \} = \texttt{LoadBalance}(\textit{V}_{1}, \cdots ,\textit{V}_{k})
\end{equation} \vspace{-3.3mm}
\begin{equation}
    \resizebox{0.91\linewidth}{!}{$
    \{\textit{Subgraph}_{i}|i \ge 1 \} = \texttt{Partition}(\textit{E}, \{\textit{Block}_{i}|i \ge 1 \})
    $}
\end{equation} 
The \textit{Block}$_{i}$ here is a set of divided nodes used to generate a \textit{Subgraph}$_{i}$. The generated sugraphs can be deployed on different devices for distributed training. Since graph partition methods are generally diverse and are widely used in a distributed system, one can easily find some classic partition methods, such as METIS \cite{METIS}, are well employed to reduce the communication cost. Herein, we discuss typical partition methods for efficient GNN training that focus on different optimization targets.

\noindent $\bullet$  \textbf{\textit{Graph partition for generic GNN training:}}
DistGNN \cite{GP_distgnn} applies a vertex-cut graph partition method to full-batch GNN training to reduce communication across partitions. The partition method preserves replicas for nodes segmented with their neighbors, which benefits the execution of delayed remote partial feature aggregation. DistDGL \cite{GP_distdgl} partitions a graph using METIS, and further co-locates nodes/edges features with graph partitions. To reduce the number of cross-partition edges and achieve a good balance, DistDGL formulates the objective as a multi-constraint problem. In this way, the original METIS is improved to solve the edge balance issue in graph partition during training.

\noindent $\bullet$  \textbf{\textit{Graph partition for sampling-based GNN training:}} sampling is becoming a time-consuming process in training large-scale graphs.
BGL \cite{GP_bgl} utilizes a graph partition method to minimize cross-partition communication in sampling subgraphs during GNN training, in which multi-source breadth first search (BFS) and greedy assignment heuristics are fully leveraged to ensure both multi-hop locality and balanced nodes distribution.
Cluster-GCN \cite{ClusterGCN} partitions a graph into multiple clusters using METIS, after which these clusters are randomly sampled to construct subgraphs for subsequent training. Cluster-GCN reduces memory usage in training and achieves a fast training speed on a deep GCN model.
Pagraph \cite{GP_Pagraph} proposes a GNN-aware graph partition method to ensure balanced partitions in workload and avoid cross-partition visits in the sampling process as much as possible. Specifically, partitioned subgraphs are extended to include \textit{L}-hop neighbors and corresponding edges for an \textit{L}-layer model, allowing independent training on each graph partition.

\subsection{Model-level Optimizations}
%%%%%%%%%%%%%%%%%%%%%%%%%%%3.2.1%%%%%%%%%%%%%%%%%%%%%%%%%%%
\subsubsection{\ding{175} GNN Simplification} 
GNN simplification is a model-specific method that simplifies operation flows in an GNN, targeting the improvement of \textbf{computation efficiency} in GNN training and inference. The layer propagation in a widely used GNN model, i.e., GCN, is given in Equation \ref{Eq1}, in which linear aggregation of neighboring information in spatial dimension and nonlinear activation are combined to update node representations. However, through recalling the design of classic yet straightforward classifiers, recent literature has argued that the efficiency of GNNs can be further promoted by simplifying redundant operations, despite the current outstanding performance. Since simplified GNNs are used in different tasks, we discuss typical models according to the model generality (or specificity).

\noindent $\bullet$  \textbf{\textit{Generic simplified GNN:}} SGC \cite{wu2019simplifying} removes the nonlinear activation, i.e., ReLU, between each layer to decrease model complexity, with only the final \texttt{Softmax} function preserved to generate probabilistic outputs. The simplified propagation is given as follows: %\vspace{-1.mm}
\begin{equation}%\vspace{-1.mm}
    \textit{Output} = \texttt{Softmax}(\textbf{S} \cdots \textbf{SSX} \textbf{W}^{1} \textbf{W}^{2} \cdots \textbf{W}^{l})
\end{equation} 
The linearized model is lightweight and includes a parameter-free part, where the initial \textbf{X} times \textbf{S} can be pre-processed without using weight matrices \textbf{W}. Leveraging above properties, SGC achieves significant acceleration on the training speed while maintaining comparable accuracy in many generic tasks, such as text and graph classification.

\noindent $\bullet$  \textbf{\textit{Special simplified GNN with tasks related:}} LightGCN \cite{he2020lightgcn} and UltraGCN \cite{mao2021ultragcn} aim to simplify the GNN model for learning embeddings from user-item interaction graphs in a recommender system. In a generic GCN layer, aggregated features are further processed via two operations: linear transformation using a learned \textbf{W} and nonlinear activation. LightGCN finds that feature transformation and nonlinear activation hardly benefit collaborative filtering by empirically exploring ablation studies on NGCF \cite{NGCF}. Therefore, LightGCN abandons the above two operations and drops self-loops in \textbf{A} to simplify the GCN model used in collaborative filtering. Moreover, UltraGCN identifies the issue in message passing of LightGCN and resolves it with a simpler structure. LightGCN passes messages by stacking multiple layers, which can cause an over-smoothing problem and is harmful to training efficiency. UltraGCN thus proposes to abandon explicit message passing among multiple layers and get approximate ultimate embeddings directly, yielding 14$\times$ speedup over LightGCN in terms of training.

%\vspace{-1.5mm}

%%%%%%%%%%%%%%%%%%%%%%%%%%%3.2.2%%%%%%%%%%%%%%%%%%%%%%%%%%%
\subsubsection{\ding{176} GNN Compression}
Current deep learning applications heavily rely on enormous data and complicated models in general. Such a model is well-representative but contains hundreds of millions of parameters, making the model training an intolerable time-consuming process. Model compression is a technique that compresses a complicated model, such as a DNN \cite{DNNAccelerteSurvey}, to a lightweight one with typically fewer parameters preserved, which is widely used to yield acceleration in training and inference and save the computation cost. By reviewing an emerging trend of applying model compression to GNNs for \textbf{efficient computation}, we discuss these methods according to their mechanisms.

\noindent $\bullet$  \textbf{\textit{Model quantification:}} as a particular quantification technique, binarization skillfully compresses model parameters and graph features in GNNs to yield significant acceleration in inference. Binarized DGCNN \cite{quantification2} and Bi-GCN \cite{quantification1} similarly introduce binarization strategies into GNNs to speed up model execution and reduce memory consumption. Degree-quant \cite{quantification3} proposes a quantization-aware training method for GNNs to enable model inference with low precision integer (INT8) arithmetic, achieving up to 4.7$\times$ speedup on CPU platform.

\noindent $\bullet$  \textbf{\textit{Knowledge distillation:}} knowledge distillation (KD) \cite{hinton2015distilling} is a technique that extracts knowledge from a teacher model and injects it into a smaller one with similar performance maintained, which at the same time, yields acceleration on model inference.
Yang et al. \cite{KD2} propose to preserve the local structure of a teacher model via a special-designed module, which helps the knowledge transfer from a trained larger model to a smaller one. Other literature \cite{KD4} presents an effective KD framework where a specially built student model can jointly benefit from a teacher model and prior knowledge. Moreover, TinyGNN \cite{KD1} bridges the gap of extracting neighbor information between a teacher mode and a student model via a combined use of a peer aware module and a neighbor distillation strategy. The learned student model can achieve dozens of times speedup in inference than the teacher model.

%################# Section 4 ##################

\section{Comparison and Analysis} \label{sec:4}

\begin{figure}[!ht] %\vspace{-3mm}
\centering
\includegraphics[trim=0 90 0 90,width=1.\columnwidth]{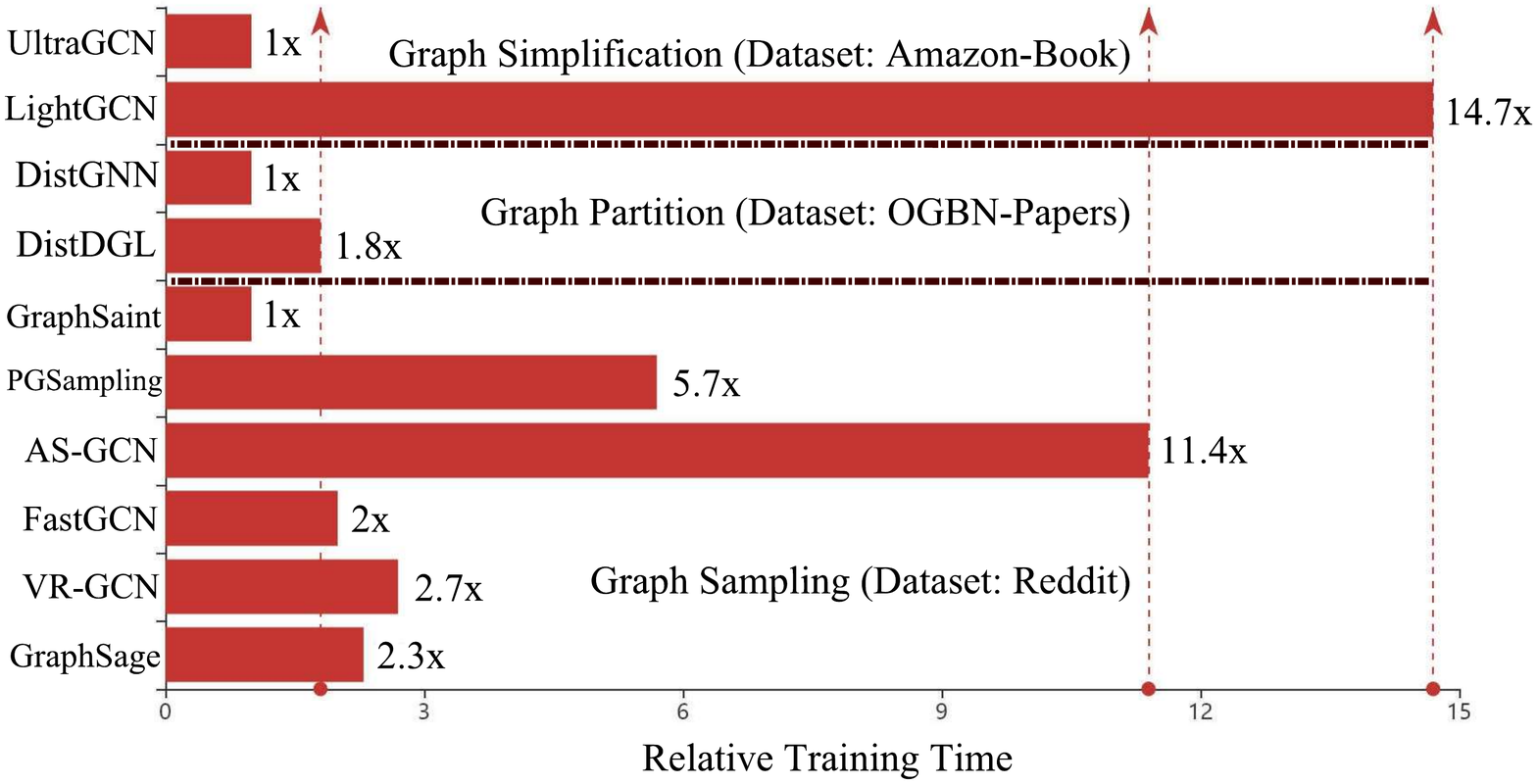}
\vspace{-6mm}
\caption{Comparison of training time among typical methods.}
\label{fig:RTT}
\end{figure}

In this section, comparison of model efficiency (reflected by relative training time) is given in Figure \ref{fig:RTT}. Note that methods in different categories are partitioned by chain lines and these in the same category are compared using the same dataset and platform. All data is collected from literature \cite{liu2021sampling,mao2021ultragcn,GP_distgnn}. We also provide overall summary and comparison of existing methods in Tables \ref{tab:Summary} \& \ref{tab:Comparision}. We pay special attention to the following aspects.

%\vspace{-2mm}

%\noindent $\bullet$  \textbf{GNN backbone} denotes which models can be applied with acceleration methods. \textit{Sampling} methods are generally conducted on spatial-based GNNs, e.g., GCNs, since neighboring connection and representation are easy to capture in a spatial dimension. A typical \textit{sparsification} method can be regarded as a \textit{sampling} method conditionally by viewing each edge as the sampling target. Moreover, \textit{sparsification} methods have wider backbones than \textit{sampling} methods for application, i.e., GCNs, GATs, and GINs. \textit{Partition} methods are mainly applied to special GNNs that are spatial-partible, e.g., GCNs and GATs, for subgraphs generation. Existing \textit{simplification} methods merely design simplified GCNs for application, owing to straightforward propagation rules and the widespread usage of GCNs. \textit{Compression} methods have been previously used for DNNs acceleration \cite{DNNAccelerteSurvey}, in which techniques that used, such as quantification, can also be deployed on GNNs and most variants with modifications.

\tabcolsep 2pt
\begin{table*}[!htb]
\centering
\begin{tabular*}{17.7cm}{cc}
\bottomrule
\textbf{Method} & \textbf{Work} \\
\bottomrule
Graph Sampling & GraphSAGE$^{1}$, VR-GCN$^{2}$, FastGCN$^{3}$, AS-GCN$^{4}$, PGSampling$^{5}$, GraphSAINT$^{6}$, HetGNN$^{7}$, HGT$^{8}$ \\
Graph Sparsification & DropEdge$^{9}$, FastGAT$^{10}$, NeuralSparse$^{11}$, SGCN$^{12}$, GAUG$^{13}$, UGS$^{14}$, AdaptiveGCN$^{15}$ \\
Graph Partition & DistGNN$^{16}$, DistDGL$^{17}$, BGL$^{18}$, Cluster-GCN$^{19}$, Pagraph$^{20}$ \\
GNN simplification & SGC$^{21}$, LightGCN$^{22}$, UltraGCN$^{23}$ \\
GNN compression & Binarized DGCNN$^{24}$, Bi-GCN$^{25}$, Degree-quant$^{26}$, KD-GCN$^{27}$, KD-framework$^{28}$, TinyGCN$^{29}$\\
\bottomrule
\end{tabular*}
\end{table*}

\tabcolsep 1.8pt
\begin{table*}[!htb]\vspace{-3.6mm}
\small{
\begin{tabular*}{17.7cm}{l}
1:\cite{hamilton2017inductive},2:\cite{chen2018stochastic},3:\cite{chen2018fastgcn},4:\cite{huang2018adaptive},5:\cite{zeng2019accurate},6:\cite{graphsaint-iclr20},\\
7:\cite{zhang2019heterogeneous},8:\cite{hu2020heterogeneous},9:\cite{dropedge1},10:\cite{sparse1},11:\cite{sparse5},12:\cite{sparse3},\\
13:\cite{dropedge2},14:\cite{sparse2},15:\cite{sparse5},16:\cite{GP_distgnn},17:\cite{GP_distdgl},18:\cite{GP_bgl},\\
19:\cite{ClusterGCN},20:\cite{GP_Pagraph},21:\cite{wu2019simplifying},22:\cite{he2020lightgcn},23:\cite{mao2021ultragcn},24:\cite{quantification2}, \\
25:\cite{quantification1},26:\cite{quantification3},27:\cite{KD2},28:\cite{KD4},29:\cite{KD1}\\
\bottomrule
\end{tabular*}}
\vspace{-2mm}
\caption{Summary and classification of the acceleration methods and corresponding work.}
\label{tab:Summary}
\end{table*}

\tabcolsep 1.8pt
\begin{table*}[!htb] %\vspace{-2.3mm}
\centering
\begin{tabular*}{17.7cm}{cccccc}
\bottomrule
\textbf{Method} & \textbf{GNN Backbone} & \textbf{Acceleration Phase} & \textbf{Optimization Objective} & \textbf{General App.} & \textbf{Special App.} \\
\bottomrule
Graph Sampling & GCN, GAT & Train. & Compt. & Node Classification & Variance Elimination \\
Graph Sparsification & GCN, GAT, GIN & Train. \& Infer. & Mem. \& Compt. & Node Classification & Denoising\\
Graph Partition & GCN, GAT & Train. & Mem. \& Com. & Node Classification & Clustering\\
GNN Simplification & GCN & Train. \& Infer. & Compt. & Node Classification & Recommendation\\
GNN Compression & GCN, GAT, GIN & Train. \& Infer. & Compt. & Node Classification & Dynamic Graphs \\
\bottomrule
\end{tabular*}
\vspace{-2mm}
\caption{Comparison among algorithmic acceleration methods from multiple aspects.}
\label{tab:Comparision}
\end{table*}

\noindent $\bullet$  \textbf{GNN backbone} denotes which models can be applied with acceleration methods. \textit{Sampling} methods are generally conducted on spatial-based GNNs, e.g., GCNs, to capture neighboring connection and representation in a spatial dimension. A typical \textit{sparsification} method can be regarded as a \textit{sampling} method conditionally by viewing each edge as the sampling target. Moreover, \textit{sparsification} methods have wider backbones than \textit{sampling} methods for application. \textit{Partition} methods are mainly applied to special GNNs that are spatial-partible for subgraphs generation. Existing \textit{simplification} methods merely design simplified GCNs for application, owing to straightforward propagation rules and the widespread usage of GCNs. \textit{Compression} methods have been previously used for DNNs acceleration \cite{DNNAccelerteSurvey}, in which techniques that used, such as quantification, can also be deployed on GNNs and most variants with modifications.

\noindent $\bullet$  \textbf{Acceleration phase} denotes which phases in GNN execution are accelerated. Since training is the most time-consuming phase, accelerating training is the primary objective for all these methods. Herein, we highlight some special cases. \textit{Sparsification} methods using learnable modules like UGS \cite{sparse2} and AdaptiveGCN \cite{sparse4} adopt sparsified graphs in inference to yield speedup. \textit{Simplification} methods generally benefit both training and inference in speed, since the cost of training and inference in a simplified model is always saved. \textit{Compression} methods provide model-level optimizations in terms of model weights and structures. Same as \textit{simplification} methods, the benefit of GNN \textit{compression} is favorable to both training and inference.

\noindent $\bullet$  \textbf{Optimization objective} denotes which objectives are optimized to yield a speedup. Most methods reduce the computation cost (abbreviated as \textbf{Compt.}) to accelerate GNN execution, such as \textit{sampling} methods. Generally, the cost of aggregating neighbor representation in a full-batch manner largely depends on the number of edges. \textit{Sparsification} methods drop useless edges in a graph to reduce memory access (abbreviated as \textbf{Mem.}) cost, thus yielding a speedup to a certain degree. \textit{Partition} methods reduce communication cost (abbreviated as \textbf{Com.}) by minimizing cross-partition edges. Moreover, memory-aware \textit{partition} methods achieve load balance according to a reasonable allocation of memory, which benefits GNN execution in terms of speed.

\noindent $\bullet$  \textbf{General and special application} denotes the general application and the special application respectively. 
Generally, all GNN based methods can resolve graph-related tasks such as (semi-supervised) node classification. Specially, \textit{sampling} methods, e.g., AS-GCN\cite{huang2018adaptive}, GraphSAINT\cite{graphsaint-iclr20}, can be leveraged to eliminate variance that introduced by probabilistic sampling. 
By regarding redundant edges in a graph as noise, \textit{sparsification} methods such as NeuralSparse\cite{sparse5} can learn a specific strategy to remove task-irrelevant edges, achieving an effect of denoising. 
Instead of randomly generating subgraphs, \textit{partition} methods can divide a graph into many smaller ones by using clustering algorithms, where size of each cluster is similar for load balance.
\textit{Simplification} methods such as LightGCN \cite{he2020lightgcn} fuse a simplified GNN models with processes such as collaborative filtering, which is designed for a recommendation task.
By adding virtual edges to graphs in a teacher model and a student model, \textit{compression} methods such as KD-GCN \cite{KD2} can extend the process of KD to dynamic graph learning.

%################## Section 5 ####################

\section{Summary and Future Prospects} \label{sec:5}

This paper provides a comprehensive survey on algorithmic acceleration methods for GNNs, in which methods in existing literature are systematically classified, discussed, and compared according to the proposed taxonomy. We believe the execution of GNNs can be promoted to gain higher efficiency via graph and model level optimizations, benefiting graph-related tasks in diverse platforms. Despite recent success and great leap of GNN acceleration methods, there are still challenges to be solved in this research field. We thereby suggest some promising prospects for future research as follows.

\noindent $\bullet$  \textbf{\textit{Acceleration for dynamic graphs:}} most acceleration methods adopt static graphs for research. A dynamic graph, however, is more flexible in topology and feature spaces than a static one, making it hard to apply these methods to dynamic graphs directly. Compression methods like KD-GCN \cite{KD2} and Binarized DGCNN \cite{quantification2} utilize a special module to extend the use to dynamic graphs, providing a nascent exemplar of dynamic graphs acceleration.

%\noindent $\bullet$  \textbf{\textit{Adjustable GNN models:}} adjustable GNN models are flexible and adaptive for diverse scenarios such as time-sensitive applications. Performing inference in a resource-limited situation generally requires to trade off the latency and accuracy, which makes an urgent demand to design an adjustable GNN model. Recent literature \cite{zhou2021accelerating} adaptively prunes the input channel to adjust GNN layers according to the inference scenarios, yielding acceleration in GNN inference.

\noindent $\bullet$  \textbf{\textit{Hardware-friendly algorithms:}} hardware-friendly algorithms benefit the model (or algorithm) execution on general platforms by leveraging hardware features. Recent literature \cite{liu2021gnnsampler} that targets to bridge the gap between graph sampling algorithms and the hardware feature, utilizes locality-aware optimizations to yield a considerable speedup in graph sampling. However, this raises the question of what characteristics should be carefully considered to design a hardware-friendly algorithm for GNN acceleration.

\noindent $\bullet$  \textbf{\textit{Algorithm and hardware co-design}}:
different to domain-specific hardware accelerators for GNNs, e.g., HyGCN \cite{yan2020hygcn} directly tailoring hardware datapath to GNNs based on the execution-semantic characterization for GNNs~\cite{yan2020characterizing}, algorithm and hardware co-design explores the design space with both algorithm and hardware awareness. 
Taking a productive co-design in a related field (i.e., graph processing) as an example, GraphDynS~\cite{yan2019graphdyns} first optimizes the execution semantic of graph traversal algorithms and then tailors its hardware datapath to the optimized execution semantic.
Similarly, in GNN acceleration, a synergy effect on optimization can be achieved by a simultaneous design of algorithm and hardware efforts in general. However, to our knowledge, there has been rare existing work on this perspective so far.

\section*{Acknowledgments}

This work was supported by the Strategic Priority Research Program of Chinese Academy of Sciences (Grant No. XDC05000000), National Natural Science Foundation of China (Grant No. 61732018 and 61872335), Austrian-Chinese Cooperative R\&D Project (FFG and CAS) (Grant No. 171111KYSB20200002), CAS Project for Young Scientists in Basic Research (Grant No. YSBR-029), and CAS Project for Youth Innovation Promotion Association.

\bibliographystyle{named}
\bibliography{ijcai22}

\end{document}